\documentclass[10pt,a4paper,conference]{IEEEtran}
\usepackage{amsmath,graphicx}
\usepackage{algorithm}
\usepackage{algpseudocode}
\usepackage{subfigure}
\usepackage{threeparttable}
\usepackage{caption}
\usepackage{array}
\usepackage{booktabs}
\usepackage{multirow}
\usepackage{amsfonts,xcolor}
\usepackage{url}
\usepackage{hyperref}
\usepackage[hyphenbreaks]{breakurl}
\usepackage{flushend,cuted}

\usepackage{cite}

\ifCLASSINFOpdf
\else
\fi
\begin{document}
%
\title{JSRT: James-Stein Regression Tree}




%
\author{\IEEEauthorblockN{Xingchun Xiang\IEEEauthorrefmark{1},
Qingtao Tang\IEEEauthorrefmark{3},
Huaixuan Zhang\IEEEauthorrefmark{1},
Tao Dai\IEEEauthorrefmark{1}\IEEEauthorrefmark{2},
Jiawei Li\IEEEauthorrefmark{1},
Shu-Tao Xia\IEEEauthorrefmark{1}\IEEEauthorrefmark{2}
}\thanks{The first and second author contributed equally to this work.}
\IEEEauthorblockA{\IEEEauthorrefmark{1}
Tsinghua Shenzhen International Graduate School, Tsinghua University,  Shenzhen, China}
\IEEEauthorblockA{\IEEEauthorrefmark{2}
PCL Research Center of Networks and Communications, Peng Cheng Laboratory, Shenzhen, China}
\IEEEauthorblockA{\IEEEauthorrefmark{3}Meituan-Dianping, Beijing, China}

Email: tsh17xxc@gmail.com, tangqingtao@meituan.com, \\zhanghuaixuan@gmail.com, daitao.edu@gmail.com, li-jw15@mails.tsinghua.edu.cn

\footnote{The first author and the second author contributed equally in this paper.}
}


\maketitle

\begin{abstract}
Regression tree (RT) has been widely used in machine learning and data mining community. Given a target data for prediction, a regression tree is first constructed based on a training dataset before making prediction for each leaf node.
 In practice, the performance of RT relies heavily on the local mean of samples from an individual node during the tree construction/prediction stage, while neglecting the global information from different nodes, which also plays an important role.
To address this issue, we propose a novel regression tree, named James-Stein Regression Tree (JSRT) by considering global information from different nodes.
 Specifically, we incorporate the global mean information based on James-Stein estimator from different nodes during the construction/predicton stage. Besides, we analyze the generalization error of our method under the mean square error (MSE) metric.
Extensive experiments on public benchmark datasets 
verify the effectiveness and efficiency of our method, and demonstrate the superiority of our method over other RT prediction methods.
\end{abstract}

\IEEEpeerreviewmaketitle

\section{Introduction}
\label{sec:intro}
{Regression trees} {(RT)} focus on the problems in which the target variables can take continuous values (typically real numbers), such as stock price \cite{patel2015predicting}, in-hospital mortality \cite{fonarow2005risk}, landslide susceptibility \cite{felicisimo2013mapping} and mining capital cost \cite{nourali2020regression}. Apart from the direct application of single regression tree, the performance can be further improved via aggregating more trees, e.g., bootstrap aggregation \cite{breiman1996bagging}, gradient boosted trees \cite{friedman2001greedy} and Xgboost \cite{chen2016xgboost}.

In single-output regression problem, the induction of regression trees can be divided into two stages, the construction stage and the prediction stage. In the construction stage, a regression tree is learned by recursively splitting the data. Finally, the samples are divided into several disjoint subsets (i.e., the leaf nodes). In the prediction stage, a value is fitted for each leaf node and the most popular prediction is the mean of sample targets in the corresponding leaf node \cite{breiman1984classification}. For decades, several methods have been proposed to replace the sample mean prediction strategy in single-output RT, e.g., linear regression \cite{quinlan1992learning}, k-Nearest Neighbor \cite{weiss1995rule}, kernel regression \cite{deng1995multiresolution} and a hybrid system of these methods \cite{torgo1997functional}. These methods make progress for partial regression tasks; however, none of them work for all regression tasks. Besides, the improvement achieved in some of these alternative methods comes with the sacrifice of the efficiency, while expeditious decision-making based on the prediction of regression trees is in need at situations like stock market. A new approach is therefore needed for stable improvement of the single-output RT prediction for all tasks while maintaining the high efficiency.

Besides, the existing estimation methods in single-output RT have one common drawback: all of them only use the training samples in a single node, which means the global data information are not well used. Typically, the input-output relation is nonlinear and varies over datasets. Thus, predictive models which only use local data information show limited generalization performance. In recent years, convolutional neural network (CNN) has swept many fields \cite{krizhevsky2012imagenet,zhang2018visual}. One main advantage of CNN is that the receptive field is increased as the network goes deeper, which means deep CNN takes advantage of global information in the whole image. In the multi-output regression problem, a regularization-based method is proposed to exploits output relatedness when estimating models at leaf nodes, which verifies the importance of global effect to improve generalization performance in regression tasks \cite{jeong2020regularization}. 

In this paper, we propose a single-output regression tree that exploits global information in all training samples, using James-Stein (JS) estimator, named James-Stein regression tree (JSRT). We first use JS estimator only in the prediction stage of RT to achieve a P-JSRT. The P-JSRT estimates values for all leaf nodes simultaneously. The estimated value of each leaf node combines local data information in each specific leaf node and global data information in all the other leaf nodes. Then, we further extend this idea to the construction stage of the regression tree. For clarity, RT with JS estimation only in the construction stages is called C-JSRT and RT with JS estimation both in the construction and prediction stages is called CP-JSRT, while JSRT is the collective name for P-JSRT, C-JSRT and CP-JSRT.

The contribution in this work involves:
\begin{itemize}
    \item the first to combine global data information and local data information in the single-output RT;
    \item introducing JS estimator to construct RT;
    \item proving that the proposed P-JSRT is uniformly better \cite{seshadri1963constructing,shao2006mathematical} than the original CART in term of Mean-Square-Error (MSE).
\end{itemize}

\section{Preliminaries}
\label{sec:Pre}

\subsection{Review of regression trees}

\textbf{Regression tree construction.} The construction process of a decision tree is known as a top-down approach. Given a data set $\mathcal{D}_n = \left\{(\boldsymbol{x}_1,y_1),(\boldsymbol{x}_2,y_2),...,(\boldsymbol{x}_n,y_n)\right\}$, $\boldsymbol{x}_i \in \mathbb{R}^d$ with features $T = \left\{t_1,t_2,...,t_d\right\}$, a regression tree chooses the best splitting feature and value pair to split into two subsets (children nodes). Each child node recursively performs this splitting procedure until the stopping condition is satisfied.

The choice of a best splitting feature and value pair in one node is usually guided by minimizing the MSE. Suppose the node to be split is $N_0$, the two children nodes after splitting are $N_1$ and $N_2$, the feature and value pair chosen to split the set is $(a, v)$. The best splitting feature and value are found by solving
\begin{equation}
    \label{eq:split}
   \min_{(a, v)}[\min_{c_1}\sum_{\boldsymbol{x}_i\in N_1}(y_i-c_1)^2+\min_{c_2}\sum_{\boldsymbol{x}_i\in N_2}(y_i-c_2)^2], 
\end{equation}
where $c_1$ and $c_2$ are the optimal output of children nodes $N_1$ and $N_2$ respectively. When each child node is considered independently, the average of sample targets is usually estimated as the optimal value: 
\begin{equation}
    \label{eq:split estimate}
    \hat{c}_1=\text{ave}(y_i|\boldsymbol{x}_i\in N_1), \\
    \hat{c}_2=\text{ave}(y_i|\boldsymbol{x}_i\in N_2)
\end{equation}


\textbf{Regression tree prediction.} In the prediction stage, the traditional regression tree takes the sample mean as the predicted value for a new ovservation, which is proved to be the maximum likelihood estimation (MLE) \cite{tang2017robust} . Given an new observation $\boldsymbol{x}$, suppose it follows a path to the $i$-th leaf node in the constructed tree. Then the predicted value $\hat{y}$ for $\boldsymbol{x}$ is $\hat{y}=\tilde{y}_i$, where $\tilde{y}_i$ is the mean of sample targets in the $i$-th leaf node which can be calculated via Eq.~(\ref{eq:split estimate}). For simplicity, we use MLE to represent the sample targets mean method in later parts of this paper.

\subsection{Related work}

Some efforts have been made to improve tree leaf approximations by fitting more complex models within each partition.

\textbf{Linear regression} is one of the most used statistical techniques and it has been introduced into regression trees to improve the prediction \cite{quinlan1992learning}. The linear regression trees have the advantage of simplicity while with lackluster performance in most cases.
\textbf{The k-nearest neighbor (KNN)} is also one of the simplest regression methods, in which only the $k$ nearest neighbors are considered when predicting the value for an unknown sample. Also, the KNN has been used in decision trees to give a more accurate classification probability \cite{weiss1995rule}.
Similarly, \textbf{kernel regressors (KR)} obtain the prediction of a new instance $\boldsymbol{x}$ by a weighted average of its neighbors. The weight of each neighbor is calculated by a function of its distance to $\boldsymbol{x}$ (called the kernel function). The kernel regression method has also been integrated with kd-trees \cite{deng1995multiresolution}.
Despite the effectiveness of KNN and KR methods, the principle of kernel regressors (i,e., involving other instances when making prediction) brings the problem of low efficiency as well.
Furthermore, a \textbf{hybrid tree learner (HTL)} was explored for better balance between accuracy and computation efficiency in regression trees \cite{torgo1997functional}. HTL applied several alternative models in the regression tree leaves, and found that kernel methods had a pretty possibility of usage in regression tree leaves. Also, the combination of different models in HTL needs careful tuning.

Apart from the unique drawback for each method stated above, they all neglect the global information in all leaf nodes, which leads to poor generalization performance in some datasets.

\subsection{Stein's phenomenon}
\label{sec:stein_phenomenon}

In decision theory and estimation theory, Stein's phenomenon is the phenomenon that when three or more parameters are estimated simultaneously, there exist combined estimators more accurate on average (that is, having lower MSE) than any method that handles the parameters separately \cite{stein1956}.

\textbf{James-Stein estimator.} JS estimator \cite{stein1956,james1961} is the best-known example of Stein's phenomenon, which behaves uniformly better than the ordinary MLE approach for the estimation of parameters in the multivariate normal distribution. Let $\mathbf{Y}$ be a normally distributed $m$-dimensional vector ($m>2$) with unknown mean $\boldsymbol{\mu}= E\mathbf{Y}$ and covariance matrix equal to the identity matrix $\boldsymbol{I}$. Given an observation $\boldsymbol{y}$, the goal is to estimate $\boldsymbol{\mu}$ by estimator $\hat{\boldsymbol{\mu}}=\phi(\boldsymbol{y})$. Denote the ordinary MLE estimator as $\hat{\boldsymbol{\mu}}_0$ and the JS estimator as $\hat{\boldsymbol{\mu}}_1$. Using the notation: $||\boldsymbol{x}||^2 = \boldsymbol{x'x}$. According to the Stein's phenomenon, the JS estimator $\hat{\boldsymbol{\mu}}_1$ leads to lower risk than the MLE estimator $\hat{\boldsymbol{\mu}}_0$ under the MSE metrics:
\begin{equation}
\label{eq:risk compare}
R(\boldsymbol{\mu}, \hat{\boldsymbol{\mu}}_1)=E||\boldsymbol{\mu} - \hat{\boldsymbol{\mu}}_1||^2<R(\boldsymbol{\mu}, \hat{\boldsymbol{\mu}}_0)=E||\boldsymbol{\mu} - \hat{\boldsymbol{\mu}}_0||^2, m \ > \ 2.
\end{equation}

\section{JSRT: James-Stein Regression Tree}
\label{sec:method}
In the RT prediction stage, we need to fit an optimal value for each of the leaf nodes. In the RT construction stage, an optimal value is also fitted to each child node when splitting a node. The existing estimation of this optimal value only considers training samples in a single node, we first propose to combine global data information from all the leaf nodes in the RT prediction stage and theoretically prove the lower generalization error of this method. Then, we extend this global information idea to the RT construction stage.

\subsection{P-JSRT: RT prediction with JS estimator}
\label{subsec:JS_prediction}
Usually, the RT independently predicts a value for each leaf node. However, Stein's phenomenon points out that if we consider the prediction of all leaf nodes as one multivariate expectation estimation problem, it is possible to achieve better results on average. Thus, we propose to exploit JS estimator to simultaneously make predictions for all the leaf nodes in RT.

Following Feldman \textit{et al}. \cite{feldman2012multi} and Shi \textit{et al}. \cite{shi2016improving}, we use the positive-part JS estimator with independent unequal variances \cite{bock1975minimax,casella1985introduction}. Specifically, the positive JS estimator is given by 
\begin{equation}
\label{eq:positive_JS}
\hat{\mu}_i^{JS+}= GM + (1-\gamma)^{+}\cdot(\tilde{y_i}-GM), m>3, \end{equation} 
where $GM=\frac{1}{m}\sum_{i=1}^m\tilde{y_i}$ is the grand mean, $\gamma=(m-3)(\sum_{i=1}^m\frac{n_i}{\sigma_i^2}(\tilde{y_i}-GM)^2)^{-1}$, and $(\cdot)^{+}=max(0,\cdot)$; $n_i$ is the number of samples in leaf node $S_i$, $\tilde{y_i}$ is the mean of sample targets in leaf node $S_i$, and $\sigma_i^2$ is the standard unbiased estimate of the variance of leaf node $S_i$.
The pseudo-code of RT prediction process with JS estimator is presented in Algorithm~\ref{alg:js prediction}.

\begin{algorithm} 	
    \caption{The prediction process in P-JSRT}
    \label{alg:js prediction} 	
    \begin{algorithmic}[1] 		
        \Require 		
        A constructed regression tree $f$.
		\Ensure 		
		A regression tree $f^{js}$, which makes prediction of the leaf nodes via JS estimator.
		\If {the leaf nodes number of $f$ is more than 3}
		    \State Make predictions for all the leaf nodes simultaneously according to the positive JS estimator in Eq. (\ref{eq:positive_JS}).
		 \Else
		    \State Use MLE to estimate the value of each leaf node independently according to Eq. (\ref{eq:split estimate}).
		 \EndIf
		\State\Return {A P-JSRT $f^{js}$.}
	\end{algorithmic} 
\end{algorithm} 

\textbf{Generalization error analysis of P-JSRT.}
In a regression problem, the model is fitted to the training data $\mathcal{D}_0(X,Y)$, which is a sample set of the true data distribution $\mathcal{D}$. Denote $\boldsymbol{x}$ as an observation in $\mathcal{D}$ and $y$ as the target of $\boldsymbol{x}$. The fitted regression tree model is marked by $f$, the output of $f$ in term of $\boldsymbol{x}$ on $\mathcal{D}$ is $f(\boldsymbol{x};\mathcal{D})$. Suppose the loss is square loss, then the generalized error on average of regression tree $f$ is
\begin{equation}
\label{eq: risk tree 1}
E(f; \mathcal{D}) = E_\mathcal{D}[(f(\boldsymbol{x}; \mathcal{D}) - y)^2].
\end{equation}
Suppose the probability density of true data distribution is $p(\boldsymbol{x}, y)$, it follows from Eq. (\ref{eq: risk tree 1}) that
\begin{equation}
\label{eq: risk tree 2}
\begin{aligned}
E(f; \mathcal{D}) = \int_\mathcal{D} (f(\boldsymbol{x}; \mathcal{D}) - y)^2 p(\boldsymbol{x}, y) d\boldsymbol{x}dy.
\end{aligned}
\end{equation}
After a regression tree is learned, the feature space is divided into $m$ disjoint subspaces (i.e., $m$ leaf nodes), $S_1, S_2,..., S_m$. Thus we have
\begin{equation}
\label{eq: risk tree 3}
\begin{aligned}
E(f; \mathcal{D}) &= \sum_{i=1}^m \int_{(\boldsymbol{x}, y), \boldsymbol{x} \in S_i} (f(\boldsymbol{x}; \mathcal{D}) - y)^2 p(\boldsymbol{x}, y) d\boldsymbol{x}dy \\
&= \sum_{i=1}^m P_i \int_{(\boldsymbol{x}, y), \boldsymbol{x} \in S_i} (f(\boldsymbol{x}; \mathcal{D}) - y)^2 \frac{p(\boldsymbol{x}, y)}{P_i} d\boldsymbol{x}dy,
\end{aligned}
\end{equation}
where $P_i = \int_{(\boldsymbol{x}, y), \boldsymbol{x} \in S_i} p(\boldsymbol{x}, y) d\boldsymbol{x}dy$.

Assume that the probability of $\mathcal{D}$ is divided equally by $S_1, S_2,..., S_m$, $i.e., P_i = P, (i=1, 2,..., m)$. Denote the estimated value of $\boldsymbol{x}$ in node $S_i$ as $\hat{y}_i = f(\boldsymbol{x};\mathcal{D}), \boldsymbol{x} \in S_i$, and the probability density function of $(\boldsymbol{x}, y), \boldsymbol{x} \in S_i$ as
\begin{equation}
\label{eq:pi}
p_i(\boldsymbol{x}, y) =
\left\{
\begin{array}{ll}
\displaystyle\frac{p(\boldsymbol{x}, y)}{P_i} &, \boldsymbol{x} \in S_i \\
0 &, {\rm other}\\
\end{array}
\right.
.
\end{equation}
Then we have
\begin{equation}
\label{eq: risk tree 4}
\begin{aligned}
&E(f; \mathcal{D}) = \sum_{i=1}^m P \int_{(\boldsymbol{x}, y), \boldsymbol{x} \in S_i} (\hat{y}_i - y)^2 \frac{p(\boldsymbol{x}, y)}{P} d\boldsymbol{x}dy \\
&= P\sum_{i=1}^m \int_{(\boldsymbol{x}, y), \boldsymbol{x} \in S_i} (\hat{y}_i - y)^2 p_i(\boldsymbol{x}, y) d\boldsymbol{x}dy \\
&= P\sum_{i=1}^{m} E_{p_i(\boldsymbol{x}, y)}(y - \hat{y}_i)^2 \\
&= P\sum_{i=1}^{m} E_{p_i(\boldsymbol{x}, y)}(y - \bar{y}_i)^2 + P\sum_{i=1}^{m} E_{p_i(\boldsymbol{x}, y)}(\bar{y}_i - \hat{y}_i)^2, \\
\end{aligned}
\end{equation}
where $\bar{y}_i = E_{p_i(\boldsymbol{x}, y)}(y)$ is the true mean of $y, \boldsymbol{x} \in S_i$. The first term of the right hand side of Eq. (\ref{eq: risk tree 4}) is irrelevant to the estimated value $\hat{y}_i$. The second term of the right hand side of Eq. (\ref{eq: risk tree 4}) can be formulated as
\begin{equation}
P\sum_{i=1}^{m} E_{p_i(\boldsymbol{x}, y)}(\bar{y}_i - \hat{y}_i)^2=P\cdot E||\bar{\boldsymbol{y}} - \hat{\boldsymbol{y}}||^2.
\end{equation}
Let us suppose when $(\boldsymbol{x}, y) \sim p_i(\boldsymbol{x}, y), y_i \sim N(\mu_i, \sigma^2)$, then we can draw the conclusion that P-JSRT achieves lower generalization error than the original CART when the number of leaf nodes is more than 2 according to Eq. (\ref{eq:risk compare}).

\subsection{C-JSRT: RT construction with JS estimator}
\label{subsec: JS_construction}
Since we achieve lower generalization error simply by using JS estimator in the RT prediction stage, we explore to combine global information in each splitting step of the RT construction stage. When splitting a node $N_0$ into two children nodes $N_1$ and $N_2$ in C-JSRT, the choice of the best feature and value pair $(a, v)$ is made by solving 
\begin{equation}
    \label{eq:C-JSRT split}
   \min_{(a, v)}[\sum_{\boldsymbol{x}_i\in N_1}(y_i-c_1)^2+\sum_{\boldsymbol{x}_i\in N_2}(y_i-c_2)^2], 
\end{equation}
where $c_1$ and $c_2$ are estimated via JS estimation. Note that JS estimation is better than MLE on average only when 3 or more parameters are estimated simultaneously. To satisfy this condition, when choosing the best feature and value pair $(a, v)$ to split node $N_0$, we consider the constructed part of a RT after splitting $N_0$ as a whole constructed RT. Therefore, we can use Algorithm~\ref{alg:js prediction} to estimate values of the `leaf nodes', including $c_1$ and $c_2$ for nodes $N_1$ and $N_2$.

However, when we estimate values of children nodes according to Eq.~(\ref{eq:positive_JS}) in C-JSRT, we found it hard to change the structure of a RT. Denote the square loss reduction caused by splitting via a feature and value pair $(a, v)$ as $\Delta L$. The reason for almost no change of C-JSRT structure is that the change (caused by JS estimation instead of MLE) to $\Delta L$ for a $(a, v)$ pair is relatively small compared with the gap of $\Delta L$ between different $(a, v)$ pairs. Therefore, the solving result of Eq.~(\ref{eq:C-JSRT split}) is the same with that of Eq.~(\ref{eq:split}). 
In other words, the chosen best splitting pair $(a, v)$ is not changed via directly using JS estimator given by Eq.~(\ref{eq:positive_JS}).

\begin{algorithm*}
	\caption{The feature selection process in C-JSRT}
	\label{alg:js construction}
	\begin{algorithmic}[1]
	\Require
		Node $N_0$ to be split; Optional non-empty feature set $(A, V)$; Current leaf nodes; The number of current leaf nodes $m_{temp}$; Stopping conditions.
	\Ensure
		The best splitting feature and value pair $(a, v)_{best}$.
	\If{node $N_0$ satisfies one or more of the stopping conditions}
	    \State Mark node $N_0$ as a leaf node; \Return
	\Else
		 \State Initialize the sum of square loss in two children nodes as $L_{min}=\infty$, the best splitting feature and value pair $(a, v)_{best}$;
	\For {each feature and value pair $(a, v)$ in $(A, V)$}
	    \State Use $(a,v)$ to split node $N_0$ into two children nodes $N_1$ and $N_2$;
		\If {$m_{temp}\geq 3$}
		    \State Use the information of current leaf nodes and the scaled JS estimator given in Eq.~(\ref{eq:scaled positive_JS}) to estimate values for children nodes $N_1$ and $N_2$ simultaneously;
		\Else
		    \State Use MLE to estimate values for children nodes $N_1$ and $N_2$ according to Eq.~(\ref{eq:split estimate}) independently;
		\EndIf
		\State Calculate the sum of square loss $L_{temp}$ in two children nodes $N_1$ and $N_2$;
		\If {$L_{temp}<L_{min}$}
		    \State $L_{min} = L_{temp}$, $(a,v)_{best} = (a,v)$;
		\EndIf
	\EndFor
	\State\Return {$(a,v)_{best}$.}
		 
	\EndIf
	\end{algorithmic}
\end{algorithm*}


A phenomenon observed in experimental results of P-JSRT inspired us to solve this problem. This phenomenon will be carefully analyzed in \ref{subsec:shrinkage property}, we briefly give the conclusion here: the reduce of MSE ($\%$) is positive correlated with the weight of grand mean ($GM$) in P-JSRT. Therefore, we introduce a new scale parameter $\lambda$ to control the weight of $GM$ in C-JSRT. The scaled positive-part JS estimator used in our C-JSRT is given by
\begin{equation}
\label{eq:scaled positive_JS}
\hat{\mu}_i^{JS+}= GM + (1-\lambda\cdot\gamma)^{+}\cdot(\tilde{y_i}-GM), m>3, \end{equation} 
where the meaning of other notations are the same as Eq.~(\ref{eq:positive_JS}). 
The feature section process in our proposed C-JSRT is concluded in Algorithm~\ref{alg:js construction}.

\section{Experimental Evaluation}
\label{sec:experiment}
\subsection{Experimental setup}

\textbf{Dataset.} We use 15 UCI datasets \cite{Dua:2019} and 1 Kaggle \cite{rodolfo2018abalone} dataset in our experiments. These datasets are designed for a wide scope of tasks and vary in the number of observations and the number of features, which are adequate to evaluate the regression tree algorithm. 
Samples with missing values in their features are removed in the experiment implementation.

\textbf{Metrics.} In regression problems, MSE on test data is the most recognized generalized error, and it is adopted to evaluate performance in our experiments.

\textbf{Settings.} To avoid the influence of data manipulation, we use 10-fold cross validation and repeat 10 times in our experiments. The minimum number of samples for one node to continue splitting is set to 20 for all the datasets. Additionally, the minimum number of samples in a leaf node is set to 5.

\subsection{Comparing P-JSRT with other RT prediction methods}
We compare our method with the original CART \cite{breiman1984classification}, KNNRT \cite{weiss1995rule} and KRT \cite{deng1995multiresolution} tree models. The distance metrics used in KRT and KNNRT are both Euclidean distance. The weight of neighbors in the KR model is the reciprocal of the distance. For fair comparison, we use sklearn toolkit in the regression tree construction stage for all methods in this part.

\begin{table}[!htbp]
	\caption{MSE comparison of P-JSRT and RT with other estimation methods.}
	\label{tab:JS_KR_KNN_MSE}
	\centering
	\small
	\begin{threeparttable}
		\begin{tabular}{ p{2.30cm}<{\centering} p{0.80cm}<{\centering} p{0.9cm}<{\centering} p{0.80cm}<{\centering} p{1.05cm}<{\centering}}
			\hline
			Dataset   & CART & KNNRT & KRT  & P-JSRT\\
			\hline
			\specialrule{0em}{1pt}{1pt}
			wine		&0.5671 & 0.5741 & \underline{\textbf{0.5105}} 	& \textbf{0.5576} \\
			\hline
			\specialrule{0em}{1pt}{1pt}
			auto\_mpg		&10.80 & \textbf{10.80} & 11.05  	& \underline{\textbf{10.77}}\\
			\hline
			\specialrule{0em}{1pt}{1pt}
			airfoil	& 	10.89 &  \underline{\textbf{8.66}} &  11.50	& \textbf{10.86}\\
			\hline
			\specialrule{0em}{1pt}{1pt}
			bike	& 	36.02 & \textbf{18.00}   &\underline{\textbf{15.15}} & 	\textbf{36.02}\\
			\hline
			\specialrule{0em}{1pt}{1pt}
			energy	& 	3.4461 & 3.4947 &\underline{\textbf{3.1263}} & 	\textbf{3.4451} \\
			\hline
			\specialrule{0em}{1pt}{1pt}
			concrete	&  21.570 & \textbf{18.144} &  \underline{\textbf{14.924}} & 	\textbf{21.556}\\
			\hline
			\specialrule{0em}{1pt}{1pt}
			compressive& 	51.55  & \textbf{46.55}  & \underline{\textbf{38.75}} 	& \textbf{51.40}\\
			\hline
			\specialrule{0em}{1pt}{1pt}
			boston	& 	19.60 &  \textbf{18.73} & \underline{\textbf{17.43}}	& \textbf{19.54}\\
			\hline
			\specialrule{0em}{1pt}{1pt}
			abalone	& 	5.9828 &  6.2288 & 6.3795	& \underline{\textbf{5.9053}}\\
			\hline
			\specialrule{0em}{1pt}{1pt}
			skill	& 	1.2758 &  1.3206 & 1.3728 	& \underline{\textbf{1.2622}}\\
			\hline
			\specialrule{0em}{1pt}{1pt}
			communities	& 0.0286 & 0.0294  & 0.0296 	& \textbf{\underline{0.0284}}\\
			\hline
			\specialrule{0em}{1pt}{1pt}
			electrical($\times 10^{-4}$) & 3.1344 & \textbf{3.0682} & \textbf{\underline{3.0171}}	& \textbf{3.1242} \\
			\hline
			\specialrule{0em}{1pt}{1pt}
			diabetes($\times 10^{3}$)	& 4.5146 & 4.5639 & 4.5907	& \underline{\textbf{4.4503}}\\
			\hline
			\specialrule{0em}{1pt}{1pt}
			pm25($\times 10^{3}$)	& 	2.4462 &  \textbf{2.2645}&  \underline{\textbf{1.9742}}	& \textbf{2.4346}\\
			\hline
			\specialrule{0em}{1pt}{1pt}
			geographical($\times 10^{3}$)& 	2.8600 &  2.8653 & \textbf{2.8466}	& \underline{\textbf{2.8065}}\\
			\hline
			\specialrule{0em}{1pt}{1pt}
			baseball($\times 10^{5}$)	& 	6.0142 & 6.3447 & 6.3126 	& \underline{\textbf{5.9761}}\\
			\hline
		\end{tabular}
		\begin{tablenotes}
			\item[*] In the first column, `electrical($\times 10^{-4}$)' means the given results of dataset electrical should multiply by $10^{-4}$. And so on, for each of the dataset after electrical in the first column.
		\end{tablenotes}
	\end{threeparttable}
\end{table}

\textbf{Effectiveness comparison.} The experimental results of MSE are displayed in Table~\ref{tab:JS_KR_KNN_MSE}. The results equal to or better than CART are highlighted in bold and the lowest MSE for each dataset is underlined. Since the experimental results about the standard deviation show no significant difference for different methods, they are not presented due to the space limitation. We notice that P-JSRT performs uniformly better \cite{shao2006mathematical} than CART, while both KNNRT and KRT methods impair the performance in about half of the datasets. Despite the relative small improvement of P-JSRT, it achieves best results in 7 datasets. Though the KNNRT and KRT methods do perform best in some datasets, they suffer from instabitily and inefficiency (this will be discussed later).

\textbf{Efficiency comparison.} Results about test time are recorded in Table~\ref{tab:JS_KR_KNN_time}, in which the datasets are placed from small to large. The results taking shortest time are highlighted in bold, and the second best results are shown in italics. It is obvious that the original CART and P-JSRT completely outperform KNNRT and KRT in term of efficiency. Besides, the prediction speed of P-JSRT is comparable to that of the original CART. Particularly, the advantage of CART and P-JSRT gradually become prominent as the size of dataset grows.

\begin{table}[!htbp]
	\caption{The test time (millisecond) of P-JSRT and RT with other estimation methods.}
	\label{tab:JS_KR_KNN_time}
	\centering
	\small
	\begin{threeparttable}
		\begin{tabular}{ p{1.35cm}<{\centering} p{1.00cm}<{\centering} p{0.58cm}<{\centering} p{0.95cm}<{\centering} p{0.97cm}<{\centering} p{1.05cm}<{\centering}}
			\hline
			Dataset   & Samples & CART & KNNRT & KRT  & P-JSRT\\
			\hline
			\specialrule{0em}{1pt}{1pt}
			concrete	&	103	&	\textbf{0.18}	&	3.22	&	3.48	&	\textit{0.32}	\\
			\hline
			baseball	&	337	&	\textbf{0.20}	&	11.34	&	12.24		&	\textit{0.34}\\
			\hline
			auto\_mpg	&	392	&	\textbf{0.19}	&	13.25	&	14.25		&	\textit{0.34}\\
			\hline
			diabetes	&	442	&	\textbf{0.20}	&	15.17	&	16.32	&	\textit{0.35}	\\
			\hline
			boston	&	503	&	\textbf{0.20}	&	18.61	&	19.93	&	\textit{0.36}	\\
			\hline
			energy	&	768	&	\textbf{0.19}	&	26.97	&	28.84	&	\textit{0.35}	\\
			\hline
			compressive	&	1030	&	\textbf{0.21}	&	43.30	&	45.78	&	\textit{0.37}	\\
			\hline
			geographical	&	1059	&	\textbf{0.22}	&	47.76	&	50.58	&	\textit{0.40}	\\
			\hline
			airfoil	&	1503	&	\textbf{0.21}	&	70.90	&	74.76		&	\textit{0.40}\\
			\hline
			communities	&	1993	&	\textbf{0.26}	&	118.01	&	123.54	&	\textit{0.47}	\\
			\hline
			skill	&	3338	&	\textbf{0.25}	&	258.29	&	267.30	&	\textit{0.51}	\\
			\hline
			abalone	&	4177	&	\textbf{0.25}	&	378.74	&	390.34	&	\textit{0.55}	\\
			\hline
			wine	&	4898	&	\textbf{0.26}	&	486.84	&	500.18	&	\textit{0.59}	\\
			\hline
			electrical	&	10000	&	\textbf{0.33}	&	1696.70	&	1722.10		&	\textit{0.85}\\
			\hline
			bike	&	17379	&	\textbf{0.42}	&	4802.90	&	4836.50	  &	\textit{1.10}\\
			\hline
			pm25	&	41758	&	\textbf{0.87}	&	32071.00	&	32155.00	&	\textit{2.51}	\\
			\hline
			
		\end{tabular}
		\begin{tablenotes}
			\item[*] The four methods are tested on the same learned tree and with same CPU E5-2680 v2 @ 2.80GHz for every dataset.
		\end{tablenotes}
	\end{threeparttable}
\end{table}

With respect to the ratio of improvement over CART against the growth of the running time, it is apparent that our P-JSRT has its unique advantages, especially in large datasets.

\subsection{The shrinkage property of JS estimator}
\label{subsec:shrinkage property}

\textbf{A toy experiment.} Simple types of maximum-likelihood and least-squares estimation procedures do not include shrinkage effects. In contrast, shrinkage is explicit in James-Stein-type inference. Our proposed P-JSRT improves the ordinary MLE tree (i.e., CART) in the leaf prediction via combining information in other leaf nodes. 

\begin{table}[!htbp] 	
\centering 	
\begin{threeparttable} 		
\caption{An example demonstrating $JSE$ shrinks $MLE$ towards the grand mean $GM$.} 		
\centering 		
\small 		
\begin{tabular}{c|  p{0.55cm}<{\centering} p{0.55cm}<{\centering}  p{0.55cm}<{\centering} p{0.55cm}<{\centering} p{0.55cm}<{\centering} p{0.55cm}<{\centering}} 			
\hline 			
Leaf\_ID & 2 & 4 & 5 & 7 & 9 & 10\\ 			
\hline 			
MLE & 21.23 & 35.29 & 29.87 & 30.57 & 44.50 & 38.89\\ 			
\hline 			
JSE & 21.41 &  35.26 &  29.93 &  30.61 &  44.33 &  38.80\\ 			
\hline 			
$|GM\tnote{*}-MLE|$ & 12.16 &  1.90 & 3.52 & 2.82 & 11.11 & 5.50\\ 			
\hline 			
$|GM\tnote{*}-JSE|$ & \bfseries11.98 & \bfseries1.87	 & \bfseries3.46 & \bfseries2.78 & \bfseries10.94 & \bfseries5.41\\ 		\hline 		
\end{tabular} 		
\begin{tablenotes} 			
\item[*] $GM$ means the grand mean of all leaf nodes on this regression tree, which is 33.39 in this specific example.
\end{tablenotes} 		
\label{Table:JS_shrinks_MLE} 	
\end{threeparttable} 
\end{table}  
We carry out a toy experiment on a small UCI dataset \cite{Dua:2019} to better understand the shrinkage effect in P-JSRT. Results in Table~\ref{Table:JS_shrinks_MLE} verify that the P-JSRT estimate ($JSE$) is closer to the value supplied by the information in other leaf nodes ($GM$) than the original CART estimate ($MLE$).

\textbf{The influence of shrink weight.}
In this part, we delve deeper into the shrinkage property of JS estimation. When $\gamma<1$, Eq.~(\ref{eq:positive_JS}) can be reformulated as
\begin{equation}
\label{eq:shrink weight}
\begin{aligned}
\hat{\mu}_i^{JS}= GM + (1-\gamma)(y_i-GM)
=(1-\gamma)y_i+\gamma{GM},
\end{aligned}
\end{equation}
where $\gamma$ means the weight of shrink direction $GM$ in the JS estimator. It is useful to know how the weight of shrink direction is related to the improvement degree of P-JSRT. Thus, we record the average weight of shrink direction in the JS estimation and the corresponding reduce on average MSE (\%) for each dataset in a 10-times 10-fold cross validation.

\begin{figure}[!htbp]
	\centering
	\includegraphics[width=0.9\linewidth]{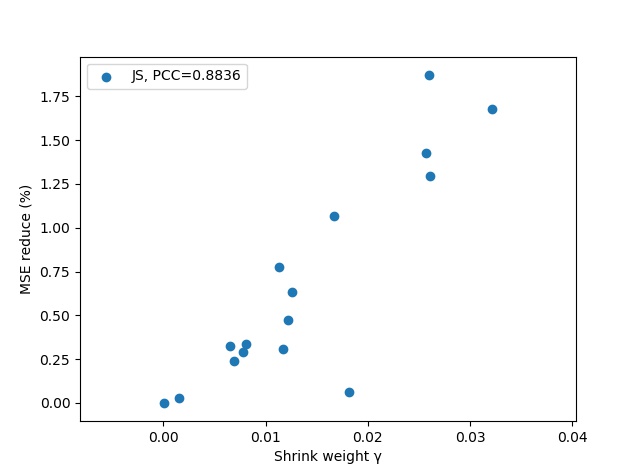}
	\caption{The influence of shrink weight $\gamma$ on the estimation. Greater the MSE reduce (\%), better the performance. PCC means the Pearson correlation coefficient between the shrink weight and the reduce on MSE (\%).}
	\label{fig:gamma_reduce}
\end{figure}

The experimental results in Fig.\ref{fig:gamma_reduce} demonstrate that the greater weight of shrink direction leads to the better performance of P-JSRT. The greater weight of shrink direction means more global information, which can avoid the overfitting problem of single regression tree to some extent. Hence, the shrinkage effect in the P-JSRT leads to lower generalization errors, greater the shrink weight, better the performance.

\begin{table}[!htbp] 	
\caption{Comparison of MSE in ablation experiment.} 	
\label{tab:ablation experiment}
\centering 	
\small 	
    \begin{threeparttable}
    \begin{tabular}{p{1.70cm}<{\centering} p{0.90cm}<{\centering} p{1.0cm}<{\centering} p{1.05cm}<{\centering} p{1.2cm}<{\centering} p{0.25cm}<{\centering}} 			
    \hline 			
    Dataset   & CART  & P-JSRT   & C-JSRT & CP-JSRT & $\lambda$\\	\hline 
    \specialrule{0em}{1pt}{1pt}
    
    concrete	&	31.7003 	&	31.6934 	&	\textbf{31.1461} 	&	\textit{31.1586} &	30		\\ 
	\hline
	baseball($\times 10^{5}$)	&	5.2823		&	5.2782	&	\textit{5.2652} 	&	\textbf{5.2622} &	25	\\
			\hline
			auto\_mpg		&	11.2340 	&	11.2266 	&	\textit{11.1963}	&	\textbf{11.1896} &	15	\\ 
			 \hline
			
			diabetes($\times 10^{3}$)	&	3.8641	&	3.8499 &			\textit{3.8238} 	&	\textbf{3.8104} &40
			\\
			\hline
			
			boston		&	26.7308 		&	\textit{26.7153} 	&	26.7290 &	\textbf{26.7136}	&	1 \\ 
			 \hline
			
			energy		&	3.3012 	&	3.3009		&	\textit{3.2929} &	\textbf{3.2926}	&	25\\ 
			\hline
			
			compressive	&	63.3120	&	63.2293 &	\textit{62.6133} 	&	\textbf{62.5370}	&	45		\\
			 \hline
			 
			airfoil		&	12.6341 	&	12.6113 	&	\textit{12.5225} &	\textbf{12.5000} &	35		\\
    \hline
	\end{tabular} 		
	\end{threeparttable} 
\end{table}

\subsection{Ablation study}
An ablation experiment is implemented to explore the efficacy of considering global information in the regression tree construction stage. We implement this ablation experiment on 8 UCI datasets. The minimum number of samples in a leaf node is set to 10 and 10-times 10-fold cross validation is used as well. We choose parameter $\lambda$ from [1, 50] with step 5. Since we have to change the construction process of the regression tree, we don't use the sklearn toolkit in this part of experiment.

The results are shown in Table~\ref{tab:ablation experiment}, the best and the second best results are highlighted in bold and italics respectively. We notice that CP-JSRT performs best in 7 of 8 datasets and the performance of C-JSRT is next to CP-JSRT, which demonstrates the efficacy of employing all training samples in both construction and prediction of the regression tree. The consistent better performance of C-JSRT/CP-JSRT than CART/P-JSRT further proves that global information should be considered in the regression tree construction stage. Besides, the fact that CP-JSRT performs better than C-JSRT and P-JSRT performs better than CART illustrates the efficacy of JS estimation in the RT prediction stage. The scale factor which controls the trade-off between local and global information during the RT construction stage varies with the dataset. 

\section{Conclusion}
\label{sec:conclusion} 
In this paper, we took advantage of global information in both the RT construction and prediction stages. Instead of obtaining the predicted value for every leaf node separately, we first introduced JS estimator to predict values for all leaf nodes simultaneously. Then, we exploit JS estimator to estimate values of children nodes in the feature selection process of RT. In this process, a trade-off between global data information and local data information was achieved. In theory, we proved that our P-JSRT had lower generalized mean square error than RT with MLE prediction under certain assumptions. Empirically, we compared the performance of P-JSRT with other estimation methods in RT, the results verified the uniform effectiveness and remarkable efficiency of our proposed P-JSRT algorithm. The ablation experiment further proved the efficacy of using JS estimation in the feature selection process of RT (method C-JSRT and CP-JSRT). In conclusion, the combination of global information and local information via JS estimator improved the generalization ability of RT.

\section*{Acknowledgment}
This work is supported in part by the National Key Research and Development Program of China under Grant 2018YFB1800204, the National Natural Science Foundation of China under Grant 61771273, the China Postdoctoral Science Foundation under Grant 2019M660645, the R\&D Program of Shenzhen under Grant JCYJ20180508152204044, and the research fund of PCL Future Regional Network Facilities for Large-scale Experiments and Applications (PCL2018KP001).





\bibliographystyle{IEEEtran} 
\bibliography{ICPR2020_xxc}

\begin{thebibliography}{10}
\providecommand{\url}[1]{#1}
\csname url@samestyle\endcsname
\providecommand{\newblock}{\relax}
\providecommand{\bibinfo}[2]{#2}
\providecommand{\BIBentrySTDinterwordspacing}{\spaceskip=0pt\relax}
\providecommand{\BIBentryALTinterwordstretchfactor}{4}
\providecommand{\BIBentryALTinterwordspacing}{\spaceskip=\fontdimen2\font plus
\BIBentryALTinterwordstretchfactor\fontdimen3\font minus
  \fontdimen4\font\relax}
\providecommand{\BIBforeignlanguage}[2]{{%
\expandafter\ifx\csname l@#1\endcsname\relax
\typeout{** WARNING: IEEEtran.bst: No hyphenation pattern has been}%
\typeout{** loaded for the language `#1'. Using the pattern for}%
\typeout{** the default language instead.}%
\else
\language=\csname l@#1\endcsname
\fi
#2}}
\providecommand{\BIBdecl}{\relax}
\BIBdecl

\bibitem{patel2015predicting}
J.~Patel, S.~Shah, P.~Thakkar \emph{et~al.}, ``Predicting stock and stock price
  index movement using trend deterministic data preparation and machine
  learning techniques,'' \emph{Expert Systems with Applications}, vol.~42,
  no.~1, pp. 259--268, 2015.

\bibitem{fonarow2005risk}
G.~C. Fonarow, K.~F. Adams, W.~T. Abraham \emph{et~al.}, ``Risk stratification
  for in-hospital mortality in acutely decompensated heart failure:
  classification and regression tree analysis,'' \emph{Jama}, vol. 293, no.~5,
  pp. 572--580, 2005.

\bibitem{felicisimo2013mapping}
{\'A}.~M. Felic{\'\i}simo, A.~Cuartero, J.~Remondo \emph{et~al.}, ``Mapping
  landslide susceptibility with logistic regression, multiple adaptive
  regression splines, classification and regression trees, and maximum entropy
  methods: a comparative study,'' \emph{Landslides}, vol.~10, no.~2, pp.
  175--189, 2013.

\bibitem{nourali2020regression}
H.~Nourali and M.~Osanloo, ``A regression-tree-based model for mining capital
  cost estimation,'' \emph{International Journal of Mining, Reclamation and
  Environment}, vol.~34, no.~2, pp. 88--100, 2020.

\bibitem{breiman1996bagging}
L.~Breiman, ``Bagging predictors,'' \emph{Machine learning}, vol.~24, no.~2,
  pp. 123--140, 1996.

\bibitem{friedman2001greedy}
J.~H. Friedman, ``Greedy function approximation: a gradient boosting machine,''
  \emph{Annals of statistics}, vol.~29, pp. 1189--1232, 2001.

\bibitem{chen2016xgboost}
T.~Chen and C.~Guestrin, ``Xgboost: A scalable tree boosting system,'' in
  \emph{Proc. of the 22nd ACM SIGKDD International Conference on Knowledge
  Discovery and Data Mining}, San Francisco, U.S.A., Aug. 2016, pp. 785--794.

\bibitem{breiman1984classification}
L.~Breiman, J.~Friedman, C.~J. Stone, and R.~A. Olshen, \emph{Classification
  and regression trees}.\hskip 1em plus 0.5em minus 0.4em\relax Boca Raton,
  U.S.A.: CRC press, 1984.

\bibitem{quinlan1992learning}
J.~R. Quinlan, ``Learning with continuous classes,'' in \emph{Proc. of the 5th
  Australian Joint Conference on Artificial Intelligence}, vol.~92, Hobart,
  Australia, Nov. 1992, pp. 343--348.

\bibitem{weiss1995rule}
S.~M. Weiss and N.~Indurkhya, ``Rule-based machine learning methods for
  functional prediction,'' \emph{Journal of Artificial Intelligence Research},
  vol.~3, pp. 383--403, 1995.

\bibitem{deng1995multiresolution}
K.~Deng and A.~W. Moore, ``Multiresolution instance-based learning,'' in
  \emph{Proc. IJCAI'95}, vol.~95, Montreal, Canada, Aug. 1995, pp. 1233--1239.

\bibitem{torgo1997functional}
L.~Torgo, ``Functional models for regression tree leaves,'' in \emph{Proc.
  ICML'97}, vol.~97, Nashville, USA, Jul. 1997, pp. 385--393.

\bibitem{krizhevsky2012imagenet}
A.~Krizhevsky, I.~Sutskever, and G.~E. Hinton, ``Imagenet classification with
  deep convolutional neural networks,'' in \emph{Advances in neural information
  processing systems}, Lake Tahoe, U.S.A., Dec. 2012, pp. 1097--1105.

\bibitem{zhang2018visual}
Q.-s. Zhang and S.-C. Zhu, ``Visual interpretability for deep learning: a
  survey,'' \emph{Frontiers of Information Technology \& Electronic
  Engineering}, vol.~19, no.~1, pp. 27--39, 2018.

\bibitem{jeong2020regularization}
J.-Y. Jeong, J.-S. Kang, and C.-H. Jun, ``Regularization-based model tree for
  multi-output regression,'' \emph{Information Sciences}, vol. 507, pp.
  240--255, 2020.

\bibitem{seshadri1963constructing}
V.~Seshadri, ``Constructing uniformly better estimators,'' \emph{Journal of the
  American Statistical Association}, vol.~58, no. 301, pp. 172--175, 1963.

\bibitem{shao2006mathematical}
J.~Shao, \emph{Mathematical statistics: exercises and solutions}.\hskip 1em
  plus 0.5em minus 0.4em\relax New York, U.S.A.: Springer Science \& Business
  Media, 2006.

\bibitem{tang2017robust}
Q.~Tang, T.~Dai, L.~Niu \emph{et~al.}, ``Robust survey aggregation with
  student-t distribution and sparse representation.'' in \emph{Proc. IJCAI'17},
  Melbourne, Australia, Aug. 2017, pp. 2829--2835.

\bibitem{stein1956}
C.~Stein, ``Inadmissibility of the usual estimator for the mean of a
  multivariate normal distribution,'' in \emph{Proc. of the Third Berkeley
  Symposium on Mathematical Statistics and Probability}, vol.~1, Berkeley,
  U.S.A., Jul. 1956, pp. 197--206.

\bibitem{james1961}
W.~James and C.~Stein, ``Estimation with quadratic loss,'' in \emph{Proc. of
  the Fourth Berkeley Symposium on Mathematical Statistics and Probability},
  vol.~1, Berkeley, U.S.A., Jun. 1961, pp. 361--379.

\bibitem{feldman2012multi}
S.~Feldman, M.~Gupta, and B.~Frigyik, ``Multi-task averaging,'' in
  \emph{Advances in Neural Information Processing Systems}, Lake Tahoe, U.S.A.,
  Dec. 2012, pp. 1169--1177.

\bibitem{shi2016improving}
T.~Shi, F.~Agostinelli, M.~Staib \emph{et~al.}, ``Improving survey aggregation
  with sparsely represented signals,'' in \emph{Proc. of the 22nd ACM SIGKDD
  International Conference on Knowledge Discovery and Data Mining}, San
  Francisco, U.S.A., Aug. 2016, pp. 1845--1854.

\bibitem{bock1975minimax}
M.~E. Bock, ``Minimax estimators of the mean of a multivariate normal
  distribution,'' \emph{The Annals of Statistics}, pp. 209--218, 1975.

\bibitem{casella1985introduction}
G.~Casella, ``An introduction to empirical bayes data analysis,'' \emph{The
  American Statistician}, vol.~39, no.~2, pp. 83--87, 1985.

\bibitem{Dua:2019}
D.~Dua and C.~Graff, ``{UCI} machine learning repository,''
  \url{http://archive.ics.uci.edu/ml}, 2017.

\bibitem{rodolfo2018abalone}
M.~Rodolfo, ``Abalone dataset,''
  \url{https://www.kaggle.com/rodolfomendes/abalone-dataset}, 2018.

\end{thebibliography}

\end{document}